\pdfoutput=1
\newcommand{\CLASSINPUTinnersidemargin}{0.75in}
\newcommand{\CLASSINPUToutersidemargin}{0.75in}
\newcommand{\CLASSINPUTtoptextmargin}{0.75in}
\newcommand{\CLASSINPUTbottomtextmargin}{0.75in}

\documentclass[10pt,conference,letterpaper,final]{IEEEtran}

\makeatletter
\def\mdseries@tt{m}
\def\mdseries@rm{m}
\makeatother

\renewcommand{\IEEEtitletopspaceextra}{0.25in}

\usepackage[T1]{fontenc}
\usepackage[utf8]{inputenc}

\usepackage[usenames,dvipsnames]{xcolor}
\usepackage{algorithmic}
\usepackage{amsmath}
\usepackage{array}
\usepackage[pdftex]{graphicx}
\usepackage{placeins}
\usepackage[obeyFinal]{todonotes}
\usepackage{url}
\usepackage{xspace}
\usepackage{libertine} %
\usepackage{tikz}
\usetikzlibrary{bending,positioning,calc,patterns,backgrounds,fit,decorations.pathreplacing}
\usepackage{ifdraft}
\usepackage{ifthen}

\usepackage{dblfloatfix}
\usepackage{paralist}

\usepackage[style=numeric-comp,sorting=none,minbibnames=3]{biblatex}
\usepackage[normalem]{ulem}

\usepackage{caption}
\DeclareCaptionFont{white}{\color{white}}
\DeclareCaptionFormat{listing}{\colorbox{gray}{\parbox{\linewidth}{#1#2#3}}}
\usepackage{listings,lstautogobble}
\usepackage[newfloat,finalizecache=false,frozencache=true]{minted}
\setminted{fontsize=\footnotesize}
\setminted{breaklines=true}
\setminted{autogobble=true}
\setminted{frame=bottomline}
\setminted{fontfamily=courier}

\usepackage{hyperref}

\newcommand\myshade{85}
\colorlet{mylinkcolor}{violet}
\colorlet{mycitecolor}{YellowOrange}
\colorlet{myurlcolor}{Aquamarine}

\hypersetup{
  linkcolor  = mylinkcolor!\myshade!black,
  citecolor  = mycitecolor!\myshade!black,
  urlcolor   = myurlcolor!\myshade!black,
  colorlinks = true,
  breaklinks = true, %
}
\usepackage{breakurl} %

\usepackage{cleveref}

\usepackage{xfrac}
\usepackage[detect-all,binary-units]{siunitx}
\usepackage{microtype}

\makeatletter
\newcounter{IEEE@bibentries}
\renewcommand\IEEEtriggeratref[1]{%
  \renewbibmacro{finentry}{%
    \stepcounter{IEEE@bibentries}%
    \ifthenelse{\equal{\value{IEEE@bibentries}}{#1}}
    {\finentry\@IEEEtriggercmd}
    {\finentry}%
  }%
}
\makeatother

\newcommand{\codeInline}[1]{\textit{#1}}

\newcommand{\BSS}[1]{BSS\nobreakdash-#1}
\newcommand{\BrainScaleS}[1]{BrainScaleS\nobreakdash-#1}

\begin{document}

\title{Extending BrainScaleS OS for \BrainScaleS2}

\author{
	\IEEEauthorblockN{%
		Eric Müller\IEEEauthorrefmark{1},
		Christian Mauch\IEEEauthorrefmark{1},
		Philipp Spilger\IEEEauthorrefmark{1},
		Oliver Julien Breitwieser\IEEEauthorrefmark{1},\\
		Johann Klähn,
		David Stöckel,
		Timo Wunderlich
		and Johannes Schemmel}\\
	\IEEEauthorblockA{%
		Kirchhoff-Institute for Physics\\
		Ruprecht-Karls-Universität Heidelberg, Germany\\
		\IEEEauthorrefmark{1}contributed equally\\
		Email: \{%
			mueller,cmauch,pspilger,obreitwi%
			\}@kip.uni-heidelberg.de
	}
}

\maketitle

\begin{abstract}
\BrainScaleS{2} is a mixed-signal accelerated neuromorphic system targeted for research in the fields of computational neuroscience and beyond-von-Neumann computing.
To augment its flexibility, the analog neural network core is accompanied by an embedded SIMD microprocessor.
The BrainScaleS Operating System (BrainScaleS OS) is a software stack designed for the user-friendly operation of the BrainScaleS architectures.
We present and walk through the software-architectural enhancements that were introduced for the \BrainScaleS{2} architecture.
Finally, using a second-version \BrainScaleS{2} prototype we demonstrate its application in an example experiment based on spike-based expectation maximization.

\end{abstract}

\ifoptionfinal{} {\tableofcontents}

\section{Introduction}

State-of-the-art neuromorphic architectures pose many requirements in terms of system control, data preprocessing, data exchange and data analysis.
In all these areas, software is involved in satisfying these requirements.
Several neuromorphic systems are directly used by individual researchers in collaborations, e.g., \cite{merolla2014million,furber2012overview,pfeil2013six,davies2018loihi}.
In addition, some systems are operated as experiment platforms providing access for external users~\cite{furber2012overview,pfeil2013six,schemmel2010iscas,davies2018loihi}.

Especially the latter calls for additional measures, such as clear and concise interfaces, resource management, runtime control and ---depending on data volumes--- ``grid-computing''-like data processing capabilities.
At the same time, usability and experiment reproducibility are crucial properties of all experiment platforms, including neuromorphic systems.

Modern software engineering techniques such as code review, continuous integration as well as continuous deployment can help to increase platform robustness and ensure experiment reproducibility.
Long-term hardware development roadmaps and experiment collaborations draw attention to platform sustainability.
Technical decisions need to be evaluated for potential future impact;
containing and reducing technical debt is a key objective during planning as well as development.
Regardless of being software-driven simulations\slash{}emulations, or being physical experiments, modern experiment setups more and more depend on these additional tools and skills in order to enable reproducible, correct and successful scientific research.

In \textcite{mueller2020bss1}, the authors already introduced \emph{BrainScaleS OS}, the Operating System for \BrainScaleS{1}.
This article describes modifications and enhancements of the \emph{BrainScaleS OS} architecture in the light of the second-generation \BSS{2} hardware generation.
The following sections introduce the hardware substrate and its envisioned exploitation model in neuroscientific modeling, machine learning and data processing in general.
\Cref{sec:methods_and_tools} introduces the methods and tools we employ.
In \cref{sec:implementation}, we discuss aspects of hybrid ---cf.\ \cref{sec:hybrid_operation}--- operation, hardware component identification and configuration as well as runtime control.
\Cref{sec:results} exemplifies the usage of the BrainScaleS Operation System on a simple experiment and describes larger experiments carried out in the past.
We close in \cref{sec:discussion} with a discussion of our work and give an overview over future developments.

\subsection{\protect\BrainScaleS{2} --- an Accelerated Mixed-Signal Neuromorphic Substrate}

\begin{figure}[tb]
	\begin{tabular}{cc}
	\includegraphics[width=.49\columnwidth]{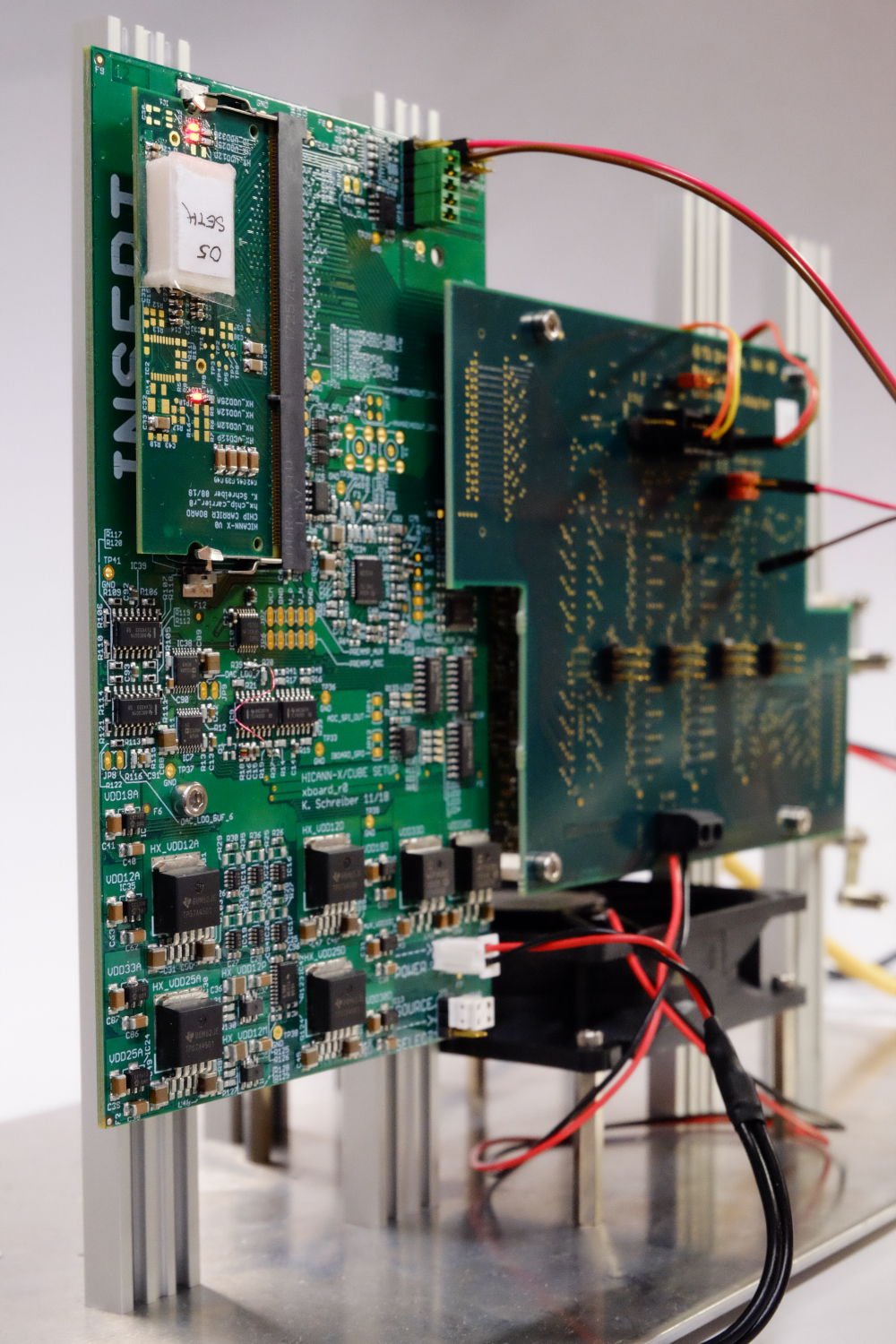}
	&
	\includegraphics[width=.44\columnwidth]{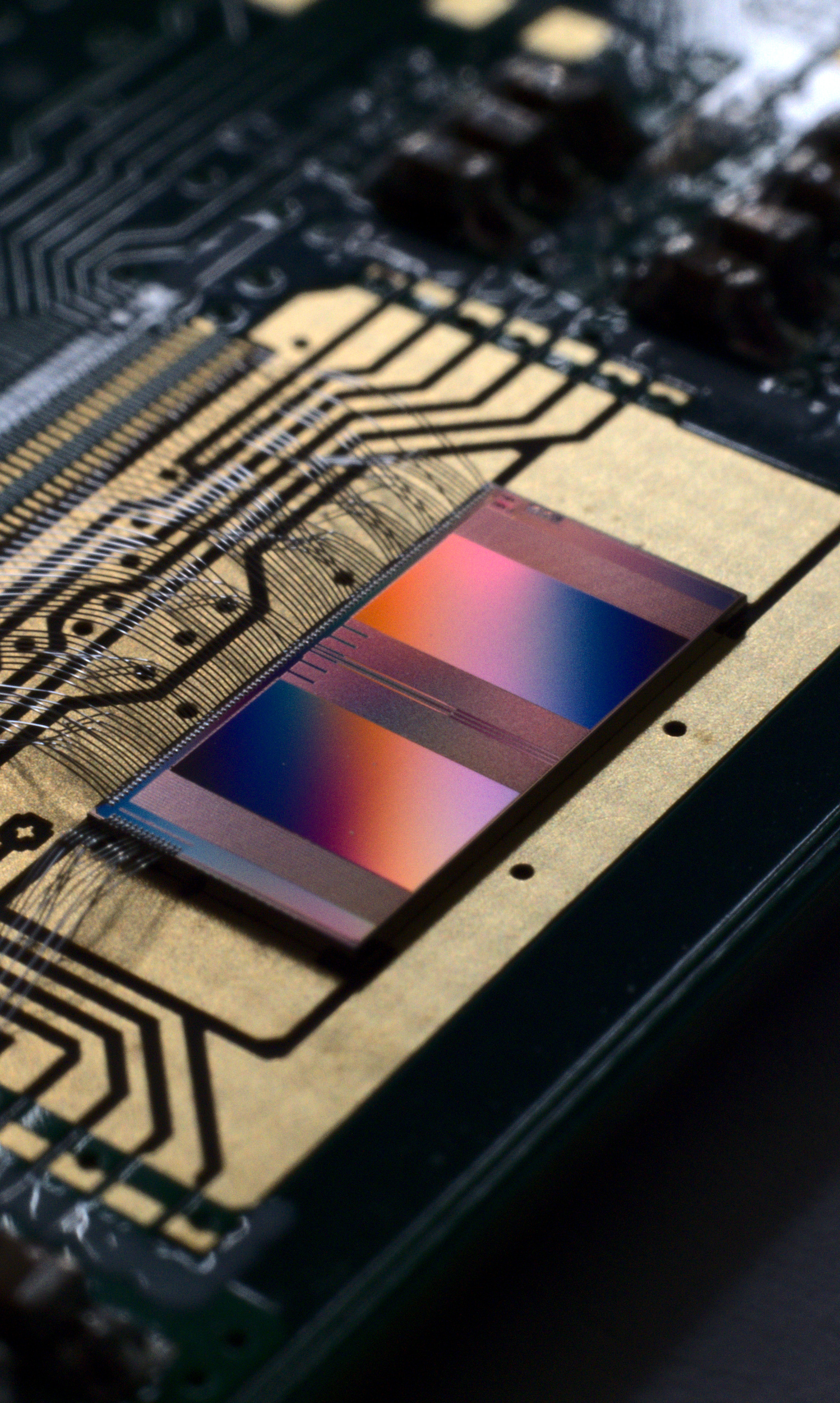}
	\end{tabular}
	\caption{\label{fig:hicannx-setup}\BrainScaleS{2} single-chip setup.
	The white plastic cap (top left) covers one accelerated neuromorphic chip (right) which is bonded onto the underlying chip-carrier PCB;
	other PCBs connect each chip to one FPGA (invisible on the back).
	The host computer and FPGAs are linked via 1-Gigabit Ethernet.
	Each \BSS{2} chip comprises $512$ AdEx neurons and $512 \times 256 = 131\,072$ synapses.
	}
\end{figure}

\Cref{fig:hicannx-setup} depicts a \BSS{2} single-chip lab setup.
The main constituent is the neuromorphic mixed-signal chip, manufactured in \SI{65}{\nano\meter} CMOS, carrying 512 AdEx neurons, $512 \times 256 = 131\,072$ plastic synapses and two embedded SIMD processors capable of fast access to the synapse matrix.
A Xilinx Kintex-7 FPGA provides the I/O interface for configuration, stimulus and recorded data.
The connection between \BSS{2} single-chip setups and a control cluster network is established via 1-Gigabit Ethernet.

The embedded SIMD microprocessor, the Plasticity Processing Unit (PPU), is a Power~\cite{powerisa_206} architecture-based single-core microprocessor with \SI{16}{\kibi\byte} SRAM.
It is equipped with a custom vector unit extension, developed in this group and designed to provide digital integer and fixed-point arithmetics which hold up with the parallelism in the analog core and access especially the synapse array in a parallel fashion~\cite{friedmann2016hybridlearning}.
In the full-sized chip each PPU features a vector unit with 128\,byte vector width, which can operate on \num{1} or \SI{2}{byte} entries.
Programs are loaded to a PPU via memory writes to the on-chip SRAM and execution is gated via a reset pin.
The PPU supports access to off-chip memory regions, e.g., the FPGA's DRAM, for instructions and scalar as well as vector data.

\subsection{Performing Experiments on Neuromorphic Systems} %

In \text{mueller2020bss1} we introduced the BrainScaleS Operating System for \BrainScaleS{1} (\BSS{1}).
It covers aspects of large-scale neuromorphic hardware configuration, experiment runtime control and platform operation.
\BSS{1} is a wafer-scale neuromorphic system that is available as an experiment platform for external researchers.
Compared to single-chip lab systems the system configuration space is large, and aspects of platform operation result in additional requirements for the software stack.
In addition, non-expert usability, operational robustness and experiment reproducibility are even more important when 
offering systems to external users.
Previous efforts~\cite{bruederle2009pyhal} focused mainly on the neuroscientific community and its view on describing spiking neural networks~\cite{davison2009pynn,eppler2008}.
However, \emph{BrainScaleS OS} has been providing access to lower-level aspects of the system to expert users~\cite{mueller2020bss1}.

We are still in the early phases of the hardware development roadmap for \BSS{2}:
the first full-sized chip arrived in the labs in 2019 after three small-chip prototypes had been produced and evaluated since 2016.
Early experiments were already implemented on the small prototypes~\cite{wunderlich2019demonstrating,billaudelle2019versatile,cramer2019control,bohnstingl2019neuromorphic}.
We make use of the same prototype version for our example experiment in \cref{sec:results}.
However, commissioning of the first full-sized \BSS{2} chip is progressing and multi-chip systems are to be expected soon.
Therefore, software requirements start to extend into regions already covered by \emph{BrainScaleS OS}.
Especially additional features in terms of structured neurons, plateau potentials and the embedded SIMD processors have to be handled by the software stack.
Another use case of \BSS{2} is its non-spiking mode resembling an analog vector-matrix multiplier that can be used in classical deep neural network experiments.

\section{Methods and Tools}\label{sec:methods_and_tools}

We already introduced functional requirements for \emph{BrainScaleS OS} enabling users to perform experiments on the \BrainScaleS{1} (\BSS{1}) neuromorphic hardware platform and additional tasks related to platform operation, cf.~\textcite{mueller2020bss1}.
In this work at hand, we focus on software aspects of configuration and control for expert usage.
We especially aim to facilitate the process of chip commissioning.
In this problem setting users need interfaces to the hardware allowing a transparent and explicit view on configuration as well as runtime control.
The implementation of the system configuration layer for \BSS{1} already provides some ideas for a structured encapsulation of configuration and runtime control.
However, while some aspects of interface are sufficient in terms of software architecture, e.g.\ the coordinate system and the strongly-typed configuration space, others needed polishing.
In particular, the description of experiment control flow was too implicit, relied on many conventions and was hard to extend or modify.
Now in \BSS{2}, the API tracks the experiment control flow explicitly, see \cref{sec:structured_configuration,sec:runtime_control}.

\subsection{Methodology and Foundations}

In~\textcite{mueller2020bss1}, we explained the design and implementation methodology: %
open-sourcing, code review, continuous integration, continuous deployment, explicit tracking of external software dependencies and containerized software development as well as user environments.

Operating custom-built experiment hardware setups poses multiple tasks:
secure and fast data exchange,
encoding/decoding of hardware configuration as well as result data,
and the definition of experiment protocols, i.e.\ a series of timed events.
Taken together, performance and correctness requirements favor the usage of a compiled language.
On the other hand, experiment description, input data preprocessing and result data analysis take advantage by interactivity and quicker turn-around cycles.

For the core software stack, we chose C++, a high-performance programming language with strong support for compile-time correctness that evolved in the last years into a multi-paradigm language.
One particular popular language for interactive usage and scripted programming is Python.
Its use in data science, computational neuroscience as well as machine learning communities enlarges the potential user base.

\subsection{Python APIs}
Exposing C++-based programming interfaces to Python can be accomplished in multiple ways.
There are at least two libraries providing support for deep integration of Python and C++, \codeInline{boost::python} and \codeInline{pybind11}~\cite{wenzel2019pybind11}.
Both libraries use advanced metaprogramming techniques to simplify the syntax and to reduce the required amount of additional code.
Among other things, aspects of type conversion, object life-time and polymorphism are handled.
However, both libraries still need some repetitive coding that could, in principle, be reduced by using a code generator operating on the C++ API's abstract syntax tree.
For the \BSS{2} software stack, we make extensive use of Clang's C++ AST accessor library, \textit{libclang}, and generate \codeInline{pybind11}-based wrapper code directly from C++ header files.
This functionality is encapsulated in the tool \codeInline{genpybind}~\cite{githubgenpybind}.

\section{Implementation}\label{sec:implementation}

\BrainScaleS{2} is a novel compute system:
on the one hand, a large fraction of the chip is dedicated to neurons, synapses and model parameter storage;
on the other hand, embedded SIMD processors enable conventional computing.
In typical neuroscientific experiments, these two parts run closely coupled.

\subsection{Hybrid Operation}\label{sec:hybrid_operation}

In this hybrid operation of the on-chip spiking neural network and the embedded SIMD processors, the latter access observables, modify parameters, perform calculations and change the input of the former to affect its dynamics.
In particular, support for flexible learning rules has been one of the main design goals for the system.

The scalar unit of the embedded SIMD processors is based on the Power instruction set architecture~\cite{powerisa_206} allowing to reuse existing open-source software infrastructure such as the C++ language infrastructure by the GNU project.
The custom vector extensions have been designed specifically for fast synaptic access.
Hence, the authors implemented support for the custom extension to facilitate its use in experiments.

\subsection{Support for the Embedded Processor}
Creating executable programs for the embedded processor ---the plasticity processor, or PPU--- is the main objective of this toolchain.
In this case, it comprises a C/C++ compiler, a linker and an assembler for our specific embedded processor.
These tools work as a cross-compilation toolchain running on a host computer and generating executables for the PPU.
The scalar part of the PPU is already supported by upstream gcc.
Extensions for the PPU's custom vector unit are described in \cref{para:toolchain}.
In addition to the core C and C++ languages, the respective standard libraries define an extensive set of additional functionality.
A subset of this functionality is appropriate for embedded programming \cite{barr1999embedded}.
The limited support for the C and C++ standard libraries is presented in \cref{para:stdlib}.
Device-specific runtime and hardware-access abstraction, beyond the general-purpose support provided via \textit{libc} and \textit{libstdc++}, is implemented in a dedicated library presented in \cref{para:libnux}.
Post-compilation and runtime supplication of parameters to PPU programs is provided via symbolic access to sections of the program using ELF (\textit{Executable and Linking Format}) information.
The supported API is presented in \cref{para:elf}.
The development of complex programs often necessitates non-trivial debugging techniques.
However, embedded systems often lack support for directly interacting with the system.
One technique addressing this problem are remote debuggers which allow debugging of problems on a different machine than on which the debugged program is running.
In \cref{para:debugger} we explain the custom remote debugger implementation for the PPU.

\subsubsection{Compiler Toolchain}\label{para:toolchain}
We use the \textit{GNU compiler collection} (gcc) together with the binary utilities package \textit{binutils} to provide this toolchain targeting \textit{C++} as programming language~\cite{gough2005introduction}.
Since the scalar part of the processor complies with a subset of the embedded Power instruction set architecture 2.06, we can take advantage of the existing gcc backend implementation.
We support the custom vector unit by providing the operation code set extension to the assembler.
Support for the PPU vector extensions was implemented similar to Power's AltiVec™~\cite{tyler1999altivec}.
Vector-unit data entities thereby become primary types on the same level as \textsf{int} with synchronization handled by the compiler transparently to the language user.
This greatly benefits the conception of plasticity algorithms as it allows, e.g., for functional and object-oriented algorithm design.

\subsubsection{C/C++ Standard Library Support}\label{para:stdlib}
The programs written for the PPU are freestanding programs.
As a consequence thereof, no system calls are available which otherwise would be provided by an operating system.
They are required for the \textit{C} and \textit{C++17}~\cite{ISO14882cxx} system libraries \textit{libc} and \textit{libstdc++} to work which are typically available in an hosted environment.
By supporting a minimal set of required system calls ---most notably page acquisition on the heap--- a slim \textit{C} library, \textit{newlib}~\cite{gatliff2002embedding}, has been integrated.
The \textit{libc} then provides the basis for \textit{libstdc++}~\cite{gcc_libstdc++} support.
Thereby full standard library support (except file system handling) is available to ease general purpose computation.
This library support can be used as a basic set of tools for implementation of re-occuring tasks in abstraction of more complex plasticity problems.
For example usage of the STL removes additional development effort of providing custom equivalent implementations.

\subsubsection{Device-specific support library}\label{para:libnux}
In order to facilitate using special features of the processor as well as the hardware, a support library has been implemented.
For instance, it provides abstracted access to a wallclock-timer, vector unit access to synapse array, a \textit{C} and \textit{C++} runtime as well as debugging functionality for stack protection facilities.
This enables reuse of frequent, at experiment runtime, or typically (e.g.,\ synapse access) needed functionality in programs implementing plasticity rules.

\subsubsection{ELF-symbol lookup functionality}\label{para:elf}
Complex plasticity kernels typically consume parameters for the initial configuration of the algorithms.
These may either be supplied at compile-time, introducing the need to recompile on parameter change, or after compilation via memory access to predefined regions.
The latter is supported by providing means to extract ELF symbol positions after compilation to the host software.
ELF (\textit{Executable and Linking Format}) \cite{tis1995elf} is a file format to store binary program data alongside with additional information, e.g., debug symbols or program section information.
By extracting section symbol name and location information, sections of the program memory layout can be annotated with symbolic names.
The user-facing API allows for \codeInline{map}-like access, e.g., \codeInline{program["my\_param"] = 24}.
This thereby allows simple symbolic access, e.g., to algorithmic parameters or to code sections, after compilation and at runtime.

\subsubsection{Debugging Support}\label{para:debugger}
An increasing level of complexity in PPU programs in conjunction with the resources limited by programming for a microprocessor with small memory constraints demands for runtime debugging capabilities.
Since the toolchain is cross-platform, i.e.\ development typically happens on a x86-based host computer while the target platform is the embedded Power-based processor, the execution in a debugger on the development platform is not possible.
However the \textit{GNU debugger} (gdb)~\cite{stallman2011debugging}, aside from normal debugging on the same machine, also offers support for remote debugging via a TCP connection to a target platform, which also works in a cross-platform setting.
The PPU being an embedded processor does neither support TCP natively nor is it feasible to implement a direct client due to memory restrictions.

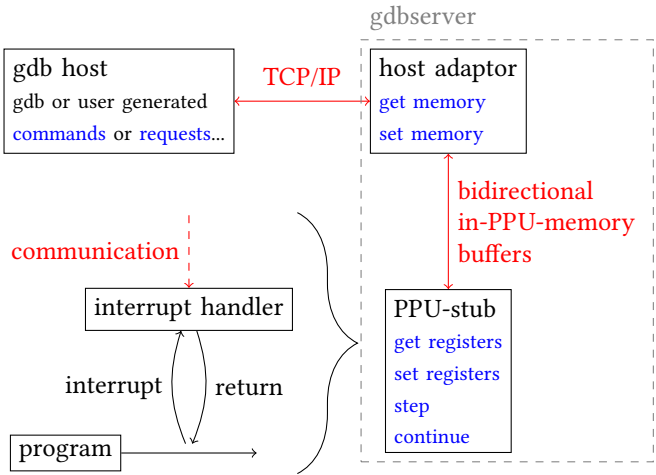
\begin{figure}[tb]
	\begin{tikzpicture}[every text node part/.style={align=left}, node distance=1.8cm]
	\node [rectangle, draw] (gdb) {gdb host\\ \footnotesize gdb or user generated \\ \footnotesize {\color{blue} commands} or {\color{blue}requests}...};
	\node [rectangle, draw, right=of gdb] (haldls) {host adaptor\\ \color{blue} \footnotesize get memory \\ \color{blue} \footnotesize set memory};
	\node [rectangle, draw, below=of haldls] (ppu) {PPU-stub\\ \color{blue}\footnotesize get registers \\ \color{blue}\footnotesize set registers \\ \color{blue}\footnotesize step \\ \color{blue}\footnotesize continue};
	\draw [<->, red, right] (haldls) -- node[midway] (com_h_p) {bidirectional \\ in-PPU-memory \\ buffers} (ppu);
	\draw [<->, red] (gdb.east) -- node[midway, above=2pt] {TCP/IP} (haldls.west);
	\node [draw, dashed, gray, fit=(haldls) (ppu) (com_h_p)] (gdbserver) {};
	\node [above right, gray] at (gdbserver.north west) {gdbserver};
	\node [left=2.3cm of ppu.south] (end) {};
	\node [left=of end,draw,rectangle] (program) {program};
	\draw [->] (program.east) -- node[midway] (mid) {} (end.west);
	\node [below=0.15cm of mid] (mid_low) {};
	\node [above=1.5cm of mid,draw,rectangle] (interrupt_handler) {interrupt handler};
	\path [->] (interrupt_handler) edge [bend left=20] node [right] {return} (mid);
	\path [->] (mid) edge [bend left=20] node [left] {interrupt} (interrupt_handler);
	\node [above=1cm of interrupt_handler] (comm) {};
	\path [<-,dashed,red] (interrupt_handler) edge node [midway,left] {\color{red}communication} (comm);
	\draw [decoration={brace,mirror,raise=41pt,amplitude=0.8cm},decorate] (mid_low) -- (comm);
\end{tikzpicture}
	\caption{\label{fig:ppu_gdb}%
	Schematic showing remote GDB debugger control flow on the embedded processor.
	A gdb instance running on the host computer communicates with remote debugging protocol via TCP/IP to the gdbserver.
	Communication from the gdbserver host adaptor to the PPU is established via in memory writes and reads.
	This allows for transporting register data and control flow information to and from the PPU program under inspection each time the interrupt handler in the PPU program is reached.
	}
\end{figure}

\Cref{fig:ppu_gdb} depicts the implementation of the \textit{gdbserver} targeting the PPU.
It is split into a minimal stub in the PPU program which understands base commands such as dumping register content to memory, replacing an instruction through a trap or stepping one instruction and a synchronous program on the host computer which communicates with the PPU through in-memory read and write operations.
This adaptor program converts requests from or produces responses to gdb via TCP.
This keeps the memory footprint in a PPU program to a minimum while allowing real-time flow control and state inspection.

\subsection{Coordinate System}\label{sec:coordinate_system}
The multitude of components on a chip leads to a large configuration space of $\approx 350\si{\kibi\byte}$.
There are over 100 distinct ranged integer and 150 boolean registers on hardware which need to be represented in software.
To provide type safety as well as other features, e.g., range checking, we do not use the builtin numeric types of \codeInline{C++} but custom ranged types\cite{githubrant}.
High symmetry in chip layout naturally leads to abstraction on different scales.

\begin{figure}[tb]
	\includegraphics[width=\columnwidth]{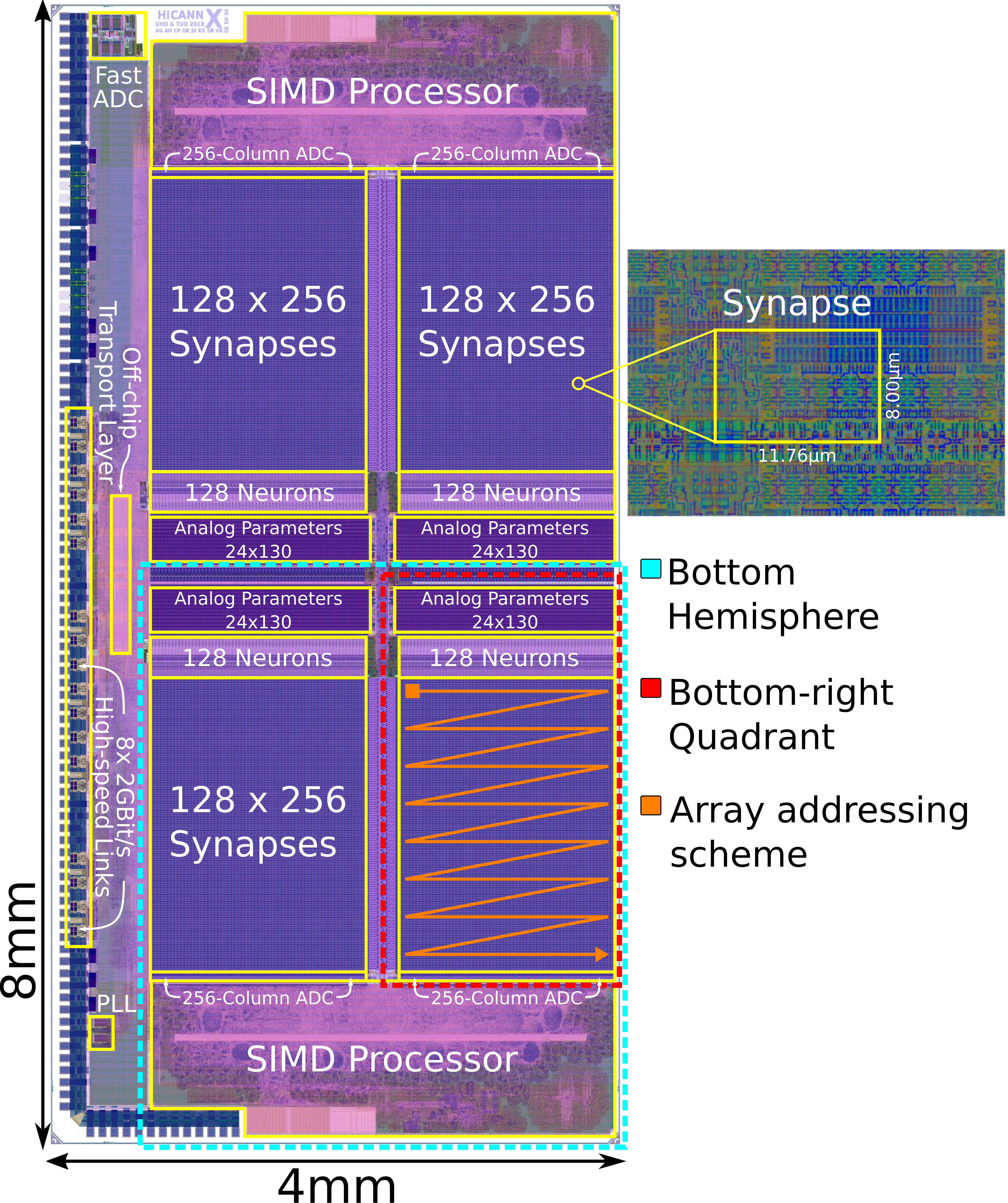}
	\caption{\label{fig:coordinates}%
	Layout schematic of the latest \BrainScaleS{2} chip.
	Various component regions are framed in yellow.
    Framed with dashed lines are logically separable regions of the chip.
    Synapses and neurons are partitioned into four quadrants, two embedded SIMD processors as well as columnar ADCs are located in the upper and lower chip hemisphere.
    The ordering scheme of two dimensional coordinates is shown in orange, rows then columns.
	}
\end{figure}

\Cref{fig:coordinates} shows the layout schematic of one chip with annotations for the different component regions.
The high symmetry of component distribution is evident, e.g., both chip hemispheres are identically and simply mirrored along the x-axis.
Some parts on the hemispheres themselves are again mirrored halves and are therefore called quadrants.
This symmetry is reflected in a hierarchical structure of the chip's coordinate system.
For \BSS{2}, the coordinate framework ---previously developed for the \emph{BrainScaleS OS}~\cite{mueller2020bss1}--- was extended.
To illustrate the idea we use the coordinate defining a Synapse circuit.
Depending on the hierarchical level, a synapse can be addressed via \codeInline{SynapseOnQuadrant}, \codeInline{SynapseOnHemisphere} or \codeInline{SynapseOnChip}.
This is helpful when implementing functionality which, for example, does not depend on which quadrant it is applied to.
Coordinates of a higher level can be cast down to a lower level, e.g.\ \codeInline{SynapseOnChip.toSynapseOnQuadrant()}.
Vice versa, lower views can be combined to create higher-level coordinates, e.g., \codeInline{SynapseOnChip(SynapseOnHemisphere, HemisphereOnChip)}.
It is also possible to convert to different components corresponding to each other.
For example, one can convert from a synapse coordinate to a neuron coordinate with \codeInline{SynapseOnChip.toNeuronOnChip()}.
As components are structured differently there is support for linear as well as two dimensional, grid-like, coordinates.
Again the synapse is an example for a grid coordinate:
it is composed of \codeInline{SynapseRowOnQuadrant} and \codeInline{SynapseColumnOnQuadrant}.

\Cref{fig:coordinates} also shows the addressing scheme (orange) which adheres to row-major order.
Furthermore, the developed ranged types enable coordinates to be used like iterators.
This facilitates, for instance, the creation of arrays with typed indexes.

\begin{listing}[tb]
    \caption{Example usage of custom coordinate type}
    \begin{minted}{c++}
for(auto synapse : iter_all<SynapseOnChip>()) {
    my_synapse_matrix[synapse].weight =
        Synapse::Weight(42);
}
    \end{minted}
    \label{listing:typed_array}
\end{listing}

\Cref{listing:typed_array} shows an example of how this is used in \codeInline{C++}.
Implementation of this coordinate system can be found at \cite{githubhalco}.

\subsection{Structuring Data for Configuration}\label{sec:structured_configuration}

To provide type-safe secure configurability of hardware entities we encapsulate the configuration in so-called containers.
A container is an object storing a representation of a possible state of a specific hardware entity or entity group.
Application of a represented state to the hardware or retrieval from the hardware is provided in a register-like fashion, the allowed operations are \codeInline{write} and \codeInline{read}.
Depending on the layer of abstraction, the granularity of access differs.
\Cref{fig:containers} shows the encapsulation at different granularities.
The implemented concepts are described in the following.

On the lowest level, access to a register-like memory location on the hardware is abstracted with the user-facing API being a configuration of a variable-length ---depending on the coordinate--- register word.
This encapsulates the state-machine behavior of a heterogeneous set of clients, e.g., SPI~\cite{hill1987serial} or Omnibus~\cite{githubomnibus}, the latter being the on-chip bus protocol featuring multi-master operation with guaranteed master-to-client ordering.
Thereby, correct usage of the underlying protocol is guaranteed, while the transported word is only restricted to the supported value range.

Building on these register-access containers, smallest-accessible continuous entities are encapsulated in containers of the so-called hardware abstraction layer (\codeInline{hal})\cite{githubhaldls}.
For the API user, the composition of sub-word configuration, corresponding to physical entities on a circuit level, are accessible as flat or hierarchical properties.
Depending on the property type, type-safe enumerations, ranged integer or boolean value types are used.
Representation of sub-word values with a one-to-one correspondence to hardware entities allow for in-code self-documenting parameter names, for example a \codeInline{NeuronConfig} might have a \codeInline{enable\_leak} boolean property or a \codeInline{refractory\_time} ranged integer value.
Named properties inherently state intent opposed to configuring raw unnamed bits of a register word manually.
A \codeInline{hal} container can encapsulate a state spanning over multiple words if the corresponding circuit or configurable entity makes use of distributed configuration bits.
On the other side, small containers may be combined, e.g., to form larger heterogeneous or repetitive containers.
In addition to type-safe access to sub-word properties, a container implements conversion to pairs of register coordinate and payload types for write operation, called \codeInline{encoding}, and register coordinates for issuing a read operation together with extraction of container state from read answer register-word payload, called \codeInline{decoding}.
To allow arbitrary grouping, the encoding and decoding for \codeInline{read} and \codeInline{write} operation is implemented using a visitor pattern to build a linear sequence of register accesses by recursively visiting sub-containers.
There may exist a $1\rightarrow{}N$ relation between a \codeInline{hal} container and multiple register container types, since a specific entity might be accessible via different communication protocols, e.g., JTAG\cite{jtagieee} and Omnibus.
The protocol is selected upon invocation of a visitor.
This allows for a unified interface for the user-facing container API and the conversion to and from register values.

\begin{figure}[tb]
	\includegraphics[width=\columnwidth]{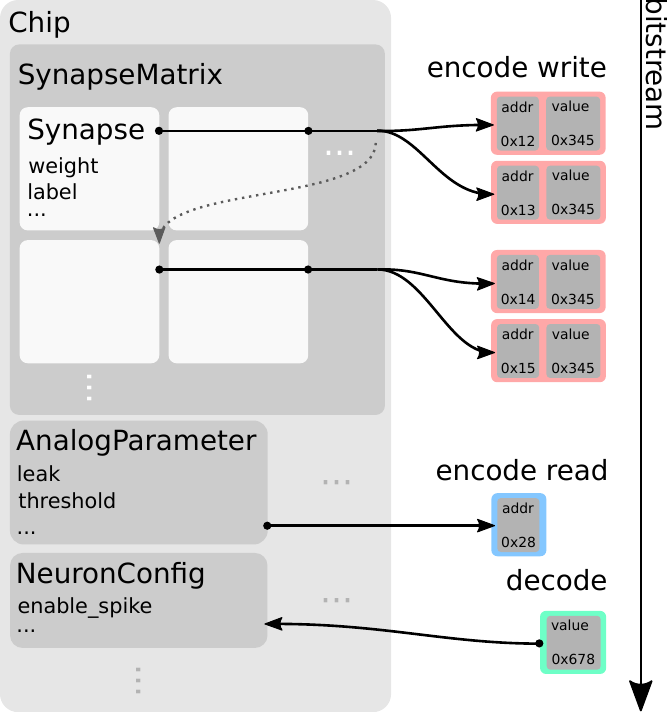}
	\caption{\label{fig:containers}
	Configurable hardware entities are modeled by nested data structures encapsulating named data elements.
	An algorithm visits the nested data structures and generates a hardware configuration bitstream.
	}
\end{figure}

\subsection{Runtime Control}\label{sec:runtime_control}

When performing experiments on a neuromorphic system, the timing of input stimulus, output data, access to observables as well as controllables is essential.
Runtime control encapsulates the time-annotated flow of how to actually use the chip.
This includes, among other things, bringing the chip into a working state, controlling of the actual spiking neuron network experiment and data transfer in general.

Multiple operational modes are supported by the hardware.
First, batch mode is suited for independent experiments.
It is characterized by not featuring read-modify-write operation via the host computer.
Conversely, the in-the-loop mode makes use of an iterative usage pattern featuring read-modify-write operations from and to an experiment controller.
It requires in-experiment synchronization.
Finally, a spiking neural network experiment that runs concurrently, time-coupled with an experiment controller is the third operation mode, the real-time closed-loop operation mode, cf.~\cref{sec:hybrid_operation}.
The controller might be located on the embedded processors or on another device.
It performs read-modify-write operations, e.g., in the form of plasticity updates or environment state variable updates within a sensor-motor loop.

To perform experiments on the neuromorphic chip, experiment descriptions need to include the sequence of timed spike events, the stimulus data.
However, the configuration of the chip might also require time-controlled execution for technical or experiment control reasons.
For instance, experiments might involve externally-triggered changes to network parameters during runtime.
In \BSS{2}, access to on-chip parameters is possible from multiple locations ---or bus masters--- the PPUs as well as the FPGA.
Hence, experiment control is distributed and the timing between these bus masters needs to be synchronized.

First, we describe the software framework for timed execution that has been developed.
Then, we illustrate the control flow of a typical experiment running on \BSS{2}.
The general concept is to construct a temporally ordered stream of commands that is sent to the communication FPGA which than handles timed release of these commands to the chip as well as time-stamped recording of responses from chip.
Constructing such a command stream, hereafter called playback program, is facilitated by three functions:
\codeInline{write}, \codeInline{read}, \codeInline{wait}.
The first two functions allow, as their names suggest, to issue write and read commands of containers, see~\cref{sec:structured_configuration}, at their respective coordinate locations.
Calling the \codeInline{read} function returns an object which provides access to the read-back data only after the experiment run, inspired by the \codeInline{std::future} class.
Write commands are likewise used to issue spike events.
Timed release of commands is facilitated by the \codeInline{wait} function allowing to delay commands relative to a timer that itself can be modified via write commands.
\Cref{listing:playback} provides a basic usage example.

\begin{listing}[tb]
    \caption{Example usage of playback builder pattern}
    \label{listing:playback}
    \begin{minted}{c++}
PlaybackProgramBuilder builder;

builder.write(NeuronConfigOnDLS(42),
    my_neuron_config);
builder.wait_until(Timer::Value(1000));
auto ticket = builder.read(SpikeCounterOnDLS(3));

auto program = builder.done();
my_executor.run(program);

auto const read_count = ticket.get();
    \end{minted}
\end{listing}

\begin{figure}[tb]
	\includegraphics[width=\columnwidth]{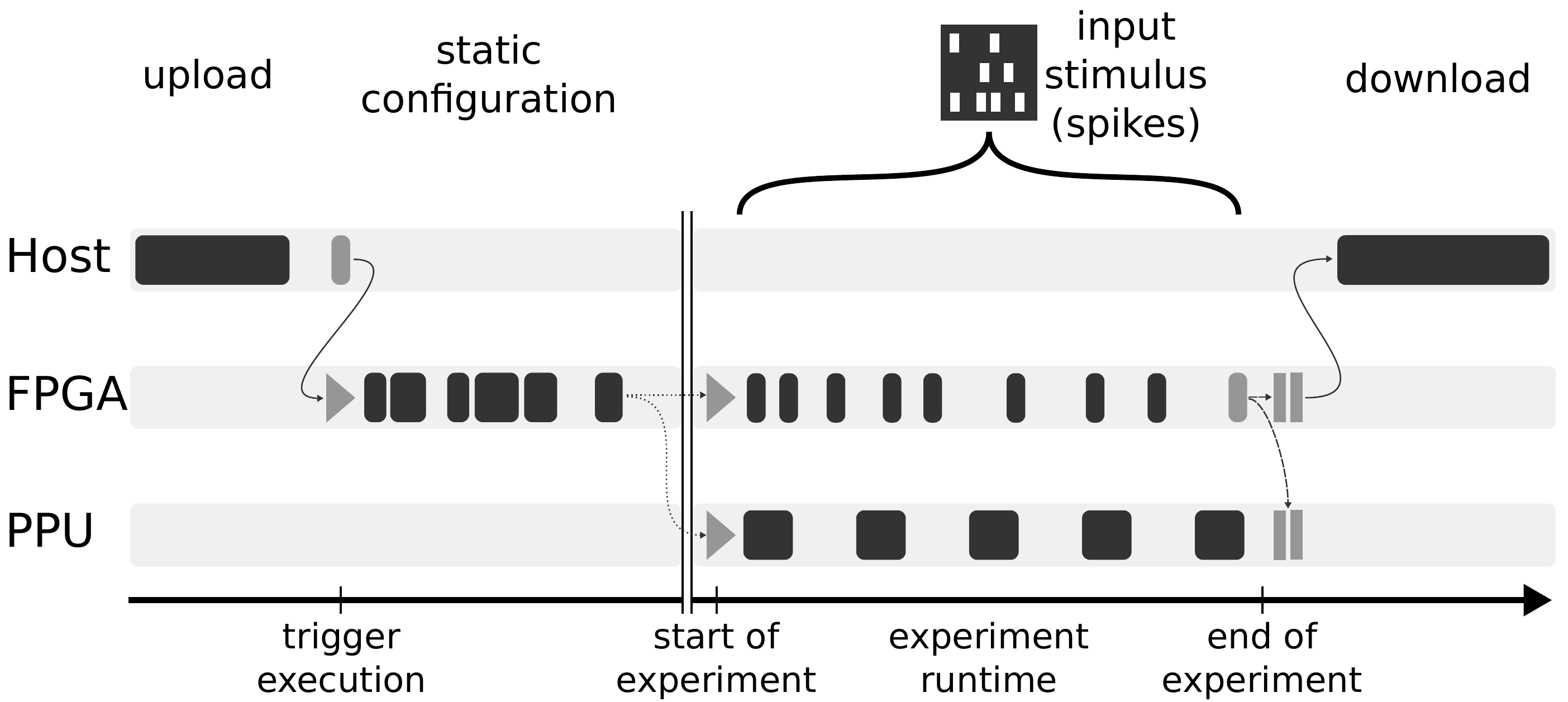}
	\caption{\label{fig:playback}Schematic showing control flow of a playback program running concurrently with code on the embedded processor.}
\end{figure}

\Cref{fig:playback} illustrates the flow of a typical experiment that involves external spike stimulus and concurrent PPU interaction.
First, communication with the FPGA is set up and subsequently used for transfer and starting of the compiled playback program.
The first stage of each program is the initial configuration of all chips components.
This stage has to be timed as analog parameters may require time to settle which makes it necessary for following commands to be delayed.
With the chip now being in a working state the experiment is started.
For this normally the timer is reset to have a zero-referenced time clock.
Likewise, execution of the PPU program is initiated.
All read responses as well as spike and ADC data are recorded with time annotation.
A special instruction defines the end of the experiment and signals the FPGA to send the recorded data to the host.

\subsection{Preemptive Experiment Scheduling}
Going one step further, this framework also allows for a separation of experiment setup, execution and analysis of hardware experiments:
instead of executing experiments on locally-attached hardware, the FPGA program is constructed on the client side, transferred to and executed at a shared remote site and its results sent back to the client.
Due to the high speed-up factor of the hardware, single experiment runtime typically ranges in the order of milliseconds real time.
Experiment assembly and result evaluation --- requiring the same order of magnitude in terms of execution time --- ordinarily happen in sequence with experiment execution.
Relegating both tasks to client side, we eliminate hardware down time.
On the remote side both experiment reception and result delivery happen in parallel to experiment execution and hence do not cause down time as well.
This allows for the hardware to be shared among several experimenters executing experiments seemingly in parallel on the same chip but more densely packed parameter sweeps for a single experimenter.
Plus, the chip remains interactively accessible as one experimenter is able to inject small experiments while a long parameter sweep is underway that would normally block anyone else from using the chip.
Overall the measures increase experiment throughput, thereby effectively speeding-up the hardware even more.

\section{Results}\label{sec:results}

Among the first experiments implemented via the hardware abstraction software framework presented in this work is the Neuromorphic Spike-based Expectation Maximization (NSEM) model.
As platform for the experiment the second \BrainScaleS{2} prototype~\cite{friedmannschemmel2016,aamir2018dls2neuron} is chosen, for which the hardware abstraction software framework presented in this work is fully implemented.

\paragraph*{Network Architecture}
As seen in \cref{fig:illu_nsem}~(top), the cause layer ---comprised of LIF-neurons brought into the stochastic regime by excitatory and inhibitory Poisson input--- receives input from an input layer that is modeled via Poisson spike trains.
Its aim is to distinguish hidden causes in the presented input stimuli.
The cause layer neurons are connected via an inhibitory population with parrot-like behavior:
each spike from a cause layer neuron elicits a spike from the inhibitory population, preventing all other cause layer neurons from firing.
The cause layer therefore forms a WTA-like structure representing a Boltzmann-machine with very strong inhibitory weights.
Therefore, it follows that only one cause layer neuron can ideally respond to each presented input pattern.
The weights $V_{ik}$ between input and cause layer evolve according to update rules~\cite{nessler2013sem,bill2015ncsem} adapted to the restrictions of the computing substrate.
The activity of each cause neuron is kept at a predetermined value via dynamic synapses, implementing a form of spike-based homeostasis heavily inspired by~\textcite{habenschuss2013stdp}.

\paragraph*{Implementation}
NSEM employs two plasticity rules (homeostasis as well as learning), acting on different synapses at different time scales.
Both are executed on the single PPU of the prototype system at the same time.
They are implemented separately and combined using a simple deadline scheduler.
In order to facilitate plasticity occurring on different timescales, each plasticity rule has a configurable deadline after which it is applied again.

\begin{figure}[htb]
  \centering%
  \includegraphics[width=\columnwidth]{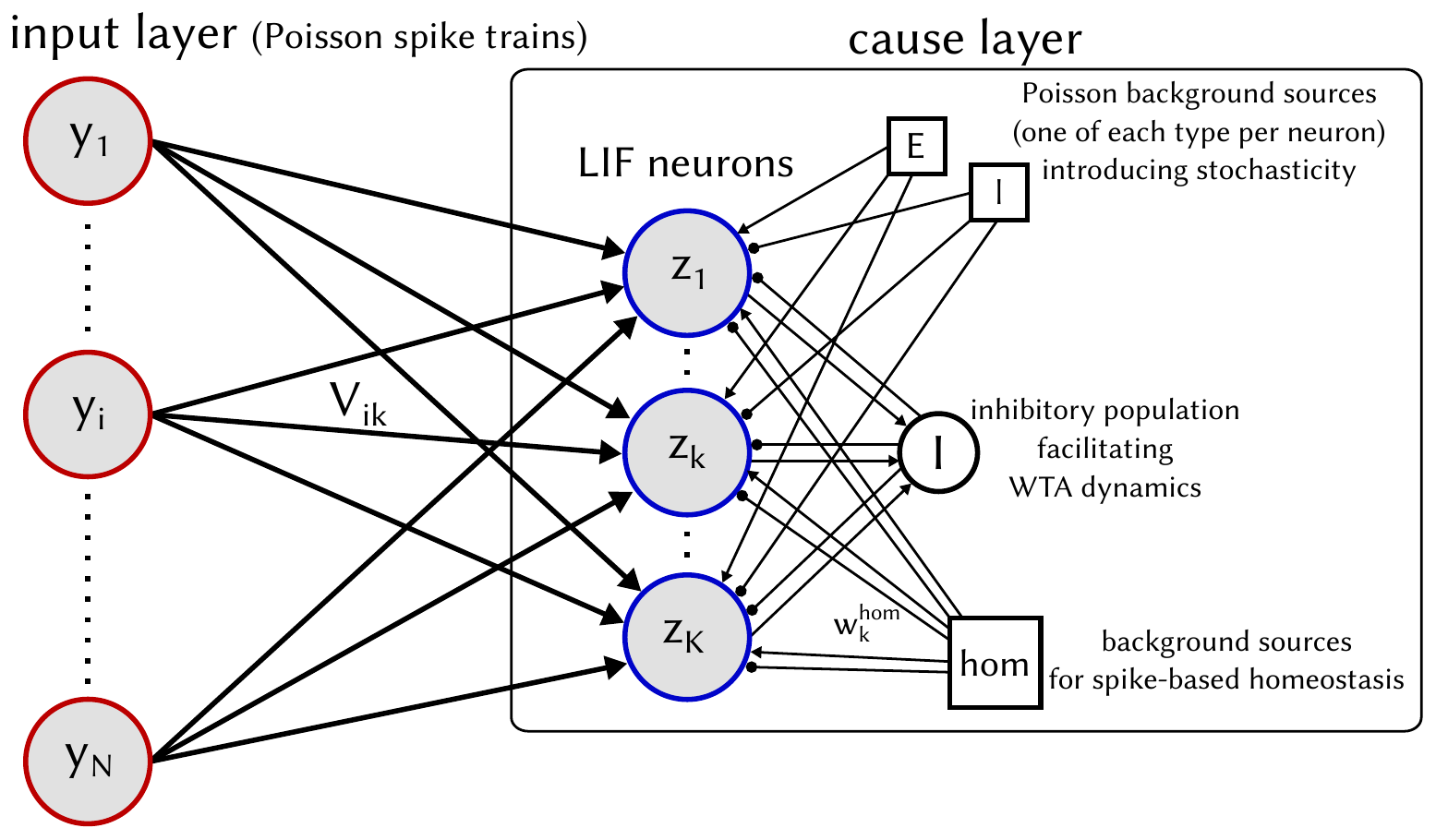}\\
  \includegraphics[width=.49\columnwidth]{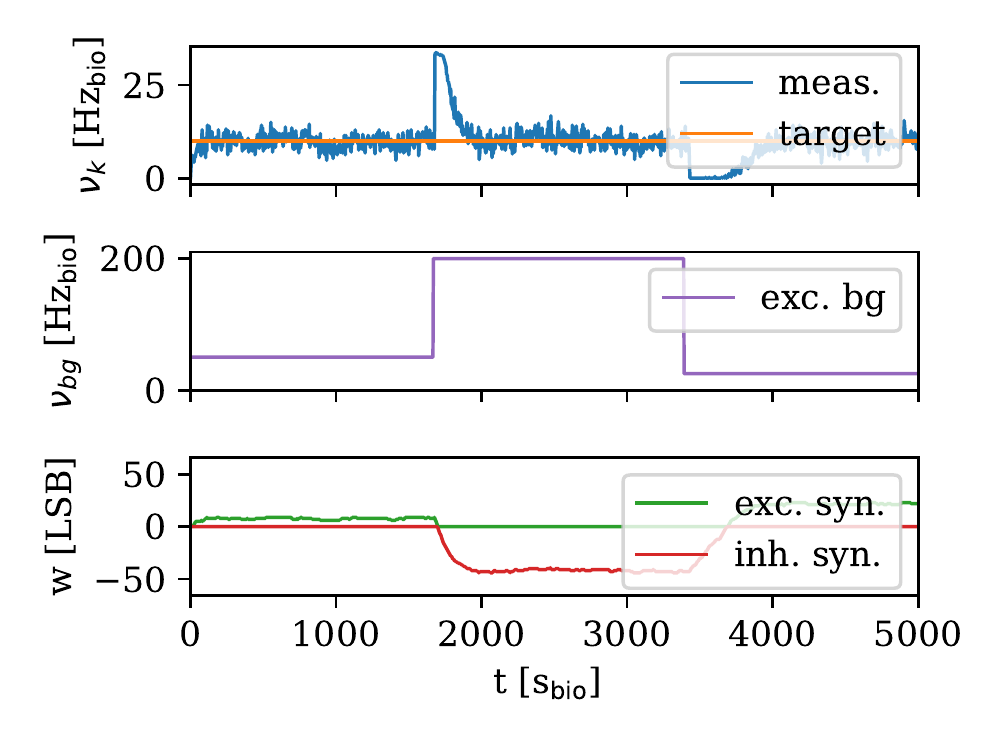}
  \includegraphics[width=.49\columnwidth]{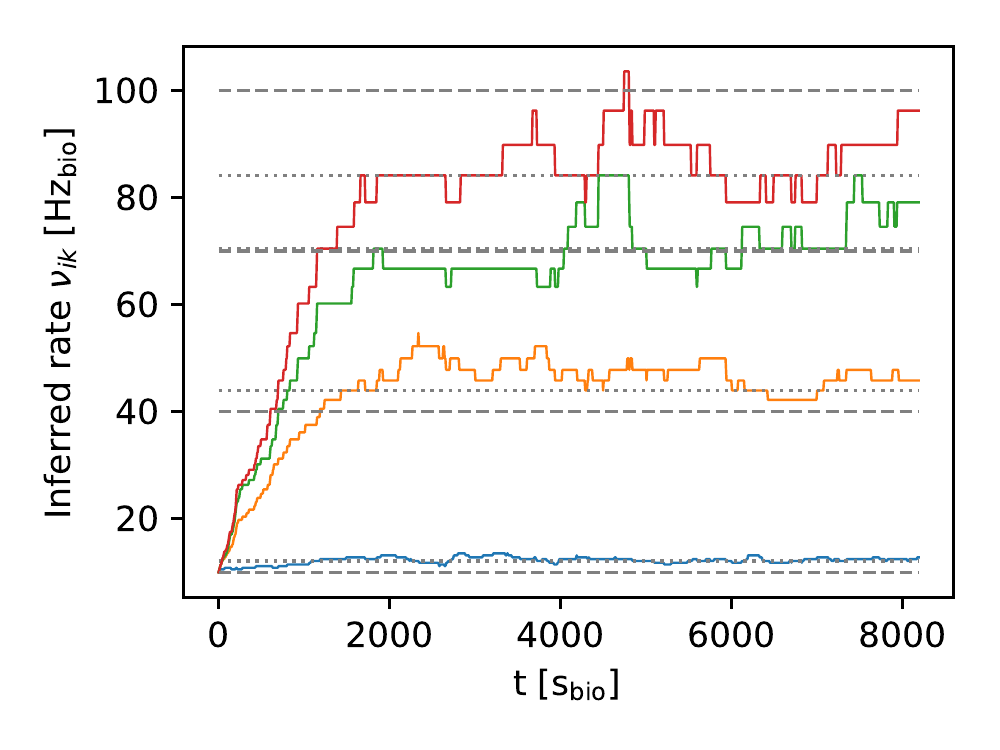}\\
  \includegraphics[width=\columnwidth]{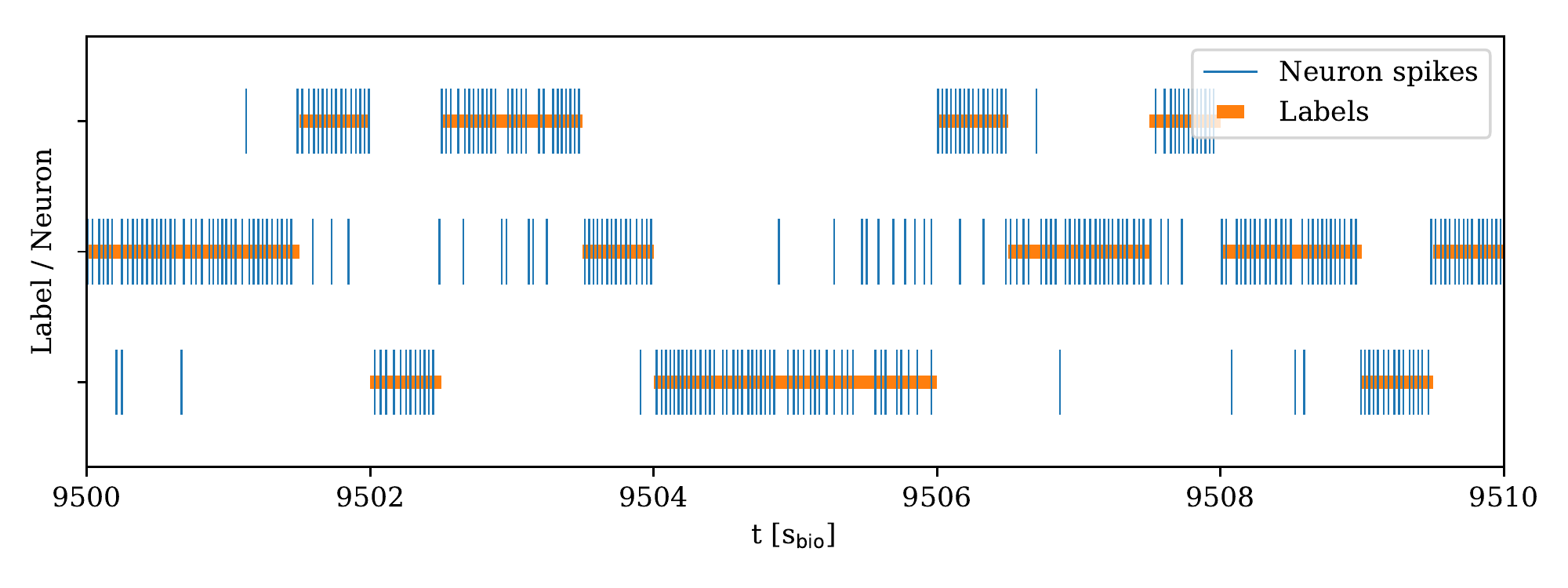}
  \caption{%
    Example experiment network architecture (top, taken from~\protect\textcite{breitwieser2015masterthesis}),
    homeostasis mechanism (center left),
    learning of target rates (center right),
    spike data as well as classification output (bottom).
	See \cref{sec:results} for more details.
  }%
  \label{fig:illu_nsem}%
\end{figure}

\paragraph*{Results}
The center sections in \cref{fig:illu_nsem} depict the different plasticity rules in action:
(center left) while a single neuron receives excitatory input from a background source with varying rate, homeostasis is able to maintain a constant firing rate after (top) sudden shifts in the background rate (middle);
the weight evolution of both homeostatic synapses (bottom) reflect the adaptive process.
(center right) Several synapses employing NSEM-rule are able to correctly infer the rate of their pre-synaptic neuron despite limited weight resolution. The dashed lines represent the expected target rate while the dotted lines represent the mean of the (colored) infered rates.
(bottom) After learning, a network of three neurons is able to differentiate between three input patterns:
for every presented pattern (bottom), a different neuron is clearly most active (top).
The network is hence able to infer hidden causes of its input in an unsupervised manner, all the while maintaining an activity equilibrium via homeostasis.
The successful implementation and execution of the experiment demonstrates the suitability for complex low-level single-chip experiments concurrently employing several distinct plasticity rules running at different time scales on the PPU.

\section{Discussion}\label{sec:discussion}

This work describes the latest developments for \emph{BrainScaleS OS} in the light of \BSS{2}.
In particular, we focus on the expert-level user interfaces for system configuration as well as experiment control.
Fundamental ideas of the software design were already devised by the authors in \textcite{bruederle2009pyhal}, the software stack for the precursor of the \BrainScaleS{1} architecture~\cite{schemmel_ijcnn06}.
Following that and due to being a large-scale neuromorphic platform~\cite{millner2010vlsi,schemmel2010iscas}, the software stack was rewritten for \BrainScaleS{1}~\cite{mueller2020bss1}.
Eventually, the second-generation BrainScaleS architecture demands for increased flexibility due to its improved configurability and programmability~\cite{aamir16dlsneuron,friedmann2016hybridlearning}.
In particular, the embedded SIMD processors require additional software during both, experiment design and experiment runtime.

\emph{BrainScaleS OS} for \BSS{2} builds upon libraries as well as development methodologies that have been developed for \BSS{1}.
In the current version, single-chip \BSS{2} systems can be robustly configured and controlled during runtime.
The software layers presented in this work focus on expert usage.
Higher-level experiment description languages such as PyNN~\cite{davison2009pynn} are not yet supported.

We demonstrate the configuration and experiment control in a hybrid experiment setting also involving
two plasticity rules, homeostasis and learning.
Multiple other experiments, e.g.~\cite{wunderlich2019demonstrating,billaudelle2019versatile,cramer2019control,bohnstingl2019neuromorphic}, demonstrate the system's applicability in machine learning and computational neuroscience.

The group also focuses on increasing reliability and quality assurance in mixed-signal hardware development~\cite{gruebill2020bss2methods}.
In particular, co-simulating software and hardware enables a continuous-integration-based workflow during hardware development.

\section{Future Developments}

\emph{BrainScaleS OS} for \BSS{2} is still under development.
The presented software layers have been sufficient for expert experimenters.
However, we identified several aspects that need attention in the future.

\subsection*{Structured Data Exchange in Distributed Systems}
Communication is a key element of distributed systems.
Inter-operation of host computers, FPGAs and PPUs typically requires communication to exchange state as well as to provide synchronization.
In particular, data is exchanged between different architectures with varying endianness and alignment constraints.
For example, exchanging plasticity parameters between host and PPU already benefits from cross-platform structured data exchange.
To solve these tasks, we aim for a thin message passing library that is integrated with a platform-agnostic serialization library.

\subsection*{Full Stack Hardware Design Validation}
Verification of hardware design prior to manufacturing is vital as chip production is expensive.
Past experiences have shown that unit testing of individual chip components alone is often insufficient.
By providing a simulator backend in a lower-level communication layer, the full software stack can be used to run tests/experiments on a simulated hardware device.
This will facilitate the validation of new FPGA features and ---most importantly--- chip designs by utilizing the complete test suite of the software stack prior to fabrication.

\subsection*{Algorithmic Task Offloading}
Increasing complexity in plasticity and first steps towards standalone on-chip calibration algorithms demand for access of arbitrary on-chip facilities.
In \cref{sec:structured_configuration} and \cref{sec:coordinate_system}, we introduced an API providing this kind of access.
Therefore, the lower-level parts of \emph{BrainScaleS OS} ---in particular the hardware abstraction layer--- are to be ported to the PPU architecture.
The availability of the vector unit for on-chip code enables optimized access to synapses and similar facilities which are unavailable for host or, more precisely, FPGA-based access.
At the same time, size and runtime performance overhead needs to be minimized.

\subsection*{Logical Experiment Description}
BrainScaleS architectures already support a variable number ---scalable to cortical connection densities--- of pre-synaptic connections per ``logical'' neuron representing multiple linked neuron circuits.
In addition, the current version of the \BSS{2} architecture makes use of a similar mechanism to build structured neurons consisting of multiple compartments;
see \textcite{aamir2018mixed} for a detailed description of the hardware implementation.
However, while the hardware abstraction layer provides support for type-safe and correct configuration of the system, it does not abstract the constitution of user-friendly encapsulation of configuration entities itself.
The set of parameters bound to such a ``logical`` configuration entity could be included in a collection of hardware abstraction layer containers possibly with constraints on their placement via coordinates.
For the encoding and decoding of these larger, logical entities, composition of hardware abstraction layer containers is to be used.
The rules under which to group parameters to ``logical`` configuration entities are currently under development.

\subsection*{Higher-level User Interfaces}
For higher-level usage, e.g., as accelerator for spiking neural network models, a topology-centric graph-based configuration API on top of the hardware abstraction established in this document is planned;
Similarly, multi-chip systems increase the need for automation of tasks related to transforming a user-defined experiment to a valid hardware configuration, hardware calibration and distributed experiment control.
Inspiration is taken from the existing higher-level software infrastructure for the \BSS{1} system~\textcite{mueller2020bss1}.

Recent developments in the machine learning community affect the way people think about data flow as well as how to programmatically describe learning algorithms~\cite{paszke2019pytorch,tensorflow2015};
on the other hand, the neuromorphic community starts building a bridge between deep neural networks and spiking neuromorphic substrates~\cite{rueckauer2017conversion,rueckauer2018conversion,goeltz2019fast}.
As a first step, the \BSS{2} non-spiking operation mode allows for a transparent integration into typical libraries for classical neural networks libraries.
Furthermore, the exploitation of the same neural network libraries allows for the specification of, e.g., plasticity rules in a computational graph;
full integration of \BSS{2} into a neural network library would be a large step towards a high-level specification of, e.g., plasticity rules.

\section{Contributions}

E.\ Müller is the lead developer and architect of the \BrainScaleS{2} software stack.
C.\ Mauch contributed to the software architecture.
P.\ Spilger contributed to the final software architecture and is the main contributor of the described experiment.
O.\ Breitwieser contributed the preemptive experiment scheduling capabilities, designed the initial experiment and contributed to the implementation on hardware.
J.\ Klähn and D.\ Stöckel contributed to the initial software architecture.
T.\ Wunderlich contributed to the remote debugger implementation.
J.\ Schemmel is the lead designer and architect of the \BrainScaleS{2} neuromorphic system.
All authors discussed and contributed to the manuscript.

\section*{Acknowledgments}

The authors wish to thank all present and former members of the Electronic Vision(s) research group contributing to the BrainScaleS-2 hardware platform, software development and operation methodologies, as well as software development.
The authors express their special gratitude towards:
\begin{inparaitem}
	\item Arthur Heimbrecht for his initial work on adding vector-unit support to the compiler;
	\item Simon Friedmann for the implementation of the initial commissioning software.
\end{inparaitem}
We especially express our gratefulness to the late Karlheinz Meier who initiated and led the project for most if its time.

This work has received funding from the EU (%
[H2020/2014-2020]%
)
under grant agreements
720270 (HBP) %
and
785907 (HBP). %

\printbibliography[notkeyword=own_software]
\printbibliography[title={Own Software},keyword=own_software]

\end{document}